\documentclass[11pt,a4paper]{article}
\usepackage[hyperref]{emnlp2020}
\usepackage{mathptmx} % <--- better than "times"
\usepackage{latexsym}
\usepackage{float}

\usepackage{microtype}

\usepackage{hyperref}
\usepackage{graphicx}
\usepackage{booktabs}
\usepackage{array}
\usepackage{listings}
\usepackage{courier}
\usepackage{enumitem}

\usepackage{titlesec}
\titlespacing*{\section}
{0pt}{0.5ex}{0ex}
\titlespacing*{\subsection}
{0pt}{0.5ex}{0ex}

\aclfinalcopy % Uncomment this line for the final submission

\newcommand{\CiteT}[1]{\citet{#1}} % When author name(s) should be read as part of the text.
\newcommand{\CiteP}[1]{~\citep{#1}} % When the citation is parenthetical (should not be read).

\newcommand{\RefSection}[1]{\hyperref[#1]{\S\,\ref{#1}}}
\newcommand{\RefFigure}[1]{\hyperref[#1]{Figure~\ref{#1}}}
\newcommand{\RefTable}[1]{\hyperref[#1]{Table~\ref{#1}}}
\newcommand{\RefListing}[1]{\hyperref[#1]{Listing~\ref{#1}}}

\newcommand{\Figure}[4]{%
\begin{figure}[H]
    \centering
    #1
    \captionsetup{
        labelfont=bf,
        justification=raggedright,
        singlelinecheck=off,
        width=.95\linewidth}
    \caption{\textbf{#3.} #4}
    \label{#2}
\end{figure}}

\newcommand{\FigureW}[4]{%
\begin{figure*}[h!]
    \centering
    #1
    \captionsetup{
        labelfont=bf,
        justification=raggedright,
        singlelinecheck=off,
        width=.95\linewidth}
    \caption{\textbf{#3.} #4}
    \label{#2}
\end{figure*}}

\newcommand{\Table}[3]{%
\begin{table}[H]
\begin{center}
  \caption{#3}
  \vspace{-0.25\baselineskip}
  \label{#2}
  #1
\end{center}
\end{table}}

\newcommand{\TableW}[3]{%
\begin{table*}[h!]
\begin{center}
  \caption{#3} 
  \vspace{-0.25\baselineskip}
  \label{#2}
  #1
\end{center}
\end{table*}}

\title{MLMLM: Link Prediction with Mean Likelihood Masked Language Model}

\author{Louis Clouatre$^\ast\dagger$, Philippe Trempe$^\ast$, Amal Zouaq$^\ast$, Sarath Chandar$^\ast\dagger$\\
$^\ast$~Polytechnique Montréal\\
$^\dagger$~\textrm{Mila}\\
\texttt{\{louis.clouatre,philippe.trempe,amal.zouaq,sarath.chandar\}@polymtl.ca}\\
\texttt{\{clouatrl,sarath.chandar\}@mila.quebec.ca}}

\begin{document}

\maketitle

\begin{abstract}
Knowledge Bases (KBs) are easy to query, verifiable, and interpretable.
They however scale with man-hours and high-quality data.
Masked Language Models (MLMs), such as BERT, scale with computing power as well as unstructured raw text data. 
The knowledge contained within those models is however not directly interpretable.
We propose to perform link prediction with MLMs to address both the KBs scalability issues and the MLMs interpretability issues.
% By committing the knowledge of MLMs to a KB, that knowledge would then become interpretable.
To do that we introduce MLMLM, \textbf{M}ean \textbf{L}ikelihood \textbf{M}asked \textbf{L}anguage \textbf{M}odel, an approach comparing the mean likelihood of generating the different entities to perform link prediction in a tractable manner.
We obtain State of the Art (SotA) results on the WN18RR dataset and the best non-entity-embedding based results on the FB15k-237 dataset.
We also obtain convincing results on link prediction on previously unseen entities, making MLMLM a suitable approach to introducing new entities to a KB. 

\end{abstract}

\section{Introduction}

\subsection{Context}

% \begin{itemize}
%     \item KB are very useful
%     \item Human interpretability/easy to query
%     \item Hard to scale
%     \item Approaches that scale with computing have proven invaluable
%     \item On the other hand, large pretrained LM scale with computing
%     \item They seem to contain world knowledge
%     \item Are not human interpretable or easy to query
% \end{itemize}

KBs have many desirable properties.
They are easy to query, verifiable, and perhaps most importantly human interpretable.
They however have one critical shortcoming, they are expensive to build making them harder to scale.
Indeed, modern KBs scale with high-quality data, manual labor, or a mix of both.
Approaches that scale with available computation and the massive amounts of unstructured data that are being created and accumulated have proven invaluable in the recent deep learning boom.

Large pretrained MLMs have been shown to scale well with large amounts of unstructured text data as well as with computing power.
They also have shown some interesting emergent abilities, such as the ability to perform zero-shot question answering \CiteP{GPT-2}.
This ability implies that the model parameters contain a large amount of factual knowledge that it can leverage to answer a wide array of questions.
That knowledge is, however, hardly interpretable by humans, as it is hidden within the hundreds of millions or even tens of billions of parameters of the language model.
% Encoder based transformer model, that currently hold the State of the Art (SotA) results in most Natural Language Understanding (NLU) benchmark do not have a straightforward mechanism to gather an answer to a query.
% For a masked language modelling approach to work, the lenght of the answer would have to be known at the time of the query.

In this paper, we are interested in exploiting MLMs for link prediction.
Many attempts at leveraging language models to complete KBs already exist.
They, however, either rely on handcrafted templates to query the model \CiteP{LMKB}, limiting the generalizability of the solution, or are intractable for any decently sized KB \CiteP{KG-BERT}.
They also generally cannot introduce new, previously unseen, entities to KB and therefore require human intervention to keep a KB up to date. 

\subsection{Motivation}

% \begin{itemize}
%     \item If LM can complete KB, solve KB scalability and LM interpretability and easy of querying
% \end{itemize}

By using MLMs to completes KBs, we can address both the issue of scalability of KBs and the issue of the interpretability of MLMs by committing knowledge of the latter to an interpretable format in the former.
The MLM can learn new knowledge from the large amount of unstructured textual data that keeps being added to the World Wide Web and then be used to continually complete and update the KB. 
This will have the very desirable effect of making the link prediction approach scale with both computational power and a large quantity of unstructured data, both of which show no sign of slowing down.

\subsection{Problem Definition}

Simply put, we want to train an MLM to, given an entity and a relation, generate all entities completing the KB triplet.

Several technical blockades had to be broken to achieve proper link prediction with pretrained MLMs.
The first one is tractability.
The models being extremely large and expensive to perform inference on, it was necessary to enable link prediction with as little inference to the model as possible.

The second one has to do with the format of the MLMs inference outputs.
The length of the output needs to be known at inference time, making it hard to sample entities of varying lengths from it.
Work like \CiteT{LMKB} is limited to single token outputs, which serves well to probe the model for the presence of embedded knowledge, but is not usable in practice for tasks such as link prediction.
Any approach has to be able to sample an MLM for entities of varying lengths to have practical applications.

Finally, the usage of MLMs opens the door to performing link prediction on unseen entities.
Some capability of such an approach with MLMs was previously demonstrated\CiteP{LMKB}.
We show that our approach yields strong results with unseen entities of arbitrary lengths in this task and should be explored further.

\subsection{Contribution}
% \subsection{Contribution}
% \begin{itemize}
%     \item development of a masked language model approach to link prediction
%     \item the first tractable MLM based link prediction approach
%     \item SotA results on one of the tested benchmark and competitive results on the other
%     \item Strong results on unseen entities
% \end{itemize}
Our main contributions are summarized here:

\setlist{nolistsep}
\begin{itemize}[noitemsep]
    \item We propose MLMLM, a mean likelihood method to compare the likelihood of different text of different token lengths sampled from an MLM.
    \item We demonstrate the tractability of our approach, which was not previously done by an MLM based model on the link prediction task.
    \item We achieve SotA results on the WN18RR benchmark and the best non entity-embedding based mean reciprocal rank on the FB15k-237 benchmark.
    \item We demonstrate that our approach can generalize reasonably well to previously unseen entities on both benchmarks.
\end{itemize}

\section{Background and Related Work}\label{sec:bg_rw}

\subsection{Masked Language Models}
% \begin{itemize}
%     \item MLM are great, BERT\CiteP{BERT} started the trend 
%     \item RoBERTa\CiteP{RoBERTa}, XLNet\CiteP{XLNet} and AlBERT\CiteP{AlBERT} are but a few of the recent models
%     \item We use the RoBERTa-Large model as we have found it to give the best performance out of those "second generation" models
%     \item Model learns to predict tokens with bi-directional context
%     \item Requires to know in advance the length of the generated chunk, we ammend this requirement (masking)
% \end{itemize}

MLMs, popularized by BERT\CiteP{BERT}, have seen tremendous success when applied to Natural Language Understanding (NLU) problems.
They are pretrained by masking tokens from text and training a large transformer encoder\CiteP{Transformer} to reconstruct the original text from the noisy inputs.
Those models incorporate enormous amounts of language knowledge and world knowledge within their weights.
This lets them be further tuned on challenging NLU tasks with great success.

Following on the footprints of BERT, several second-generation MLMs have been released.
These models\CiteP{RoBERTa, XLNet, AlBERT} have seen great improvements when compared to BERT on downstream tasks.
Among other improvements to the original training process, these models were trained for much longer with much larger text corpus to achieve those results.

Being based on the transformer encoder architecture, the output length of the model is equal to the input length.
This makes it challenging to sample text of arbitrary length when using MLMs without knowing the length of the desired sample in advance.

\subsection{Link Prediction}
% \begin{itemize}
%     \item Task of predicting what entities are in a certain relation with another
%     \item Done with ranking of all known entities based on their likelihood to be in a certain relation with another entity
%     \item Mean Reciprocal Rank, mean precision @ 1, mean precision @ 10 and Mean Rank are the most commonly used metrics
% \end{itemize}

Link prediction is the task of finding all potential entities that are in a specific relation with another entity.
A knowledge graph (KG) is composed of a set of entities $E$, a set of relations $R$, and a set of valid triplets $(h, r, t)$ representing the head entity $h$, the relation $r$ and the tail entity $t$.
By assigning a score to all possible triplets completing $(h, r, ?)$ and $(?, r, t)$, it is possible to rank all possible entities and thus complete the missing links within a KG.

\subsection{A Re-evaluation of Knowledge Graph Completion Methods}
% TODO This can be explained more, it is very important that I justify clearly why we are not using the other reported numbers
% \begin{itemize}
%     \item Recently, \CiteT{ReevaluationKB} showed that many NN based models suffered from issues during evaluation
%     \item All experiments are done with their porposed inference method
%     % \item Our proposed approach does not suffer from this issue and we used their proposed inference method for all reported results.
% \end{itemize}

Recently, \CiteT{ReevaluationKB} has found that many of the SotA approaches to link prediction have used an inappropriate evaluation protocol.
They have shown that the evaluation protocol typically used in the link prediction approaches assigns a perfect score to a constant output, by putting the correct entities on top during a tiebreaker.
In essence, under this evaluation protocol, assigning a likelihood of 0 to all entities would yield a perfect reranking score, since the tiebreaker would put the target entity as the first prediction.
This was shown to yield very inflated scores for many neural network based link prediction approaches \CiteP{NN1, NN2, NN3}, as several of them output a large number of tied scores for the various entities.
Entity embedding-based approaches \CiteP{EMB1, EMB2, EMB3} do not suffer from this issue.
While we have found that our approach does not suffer from this issue despite not being an embedding approach, we will use the random evaluation protocol proposed by \CiteT{ReevaluationKB} for all evaluations and compare against approaches that used a similar protocol to ensure the validity of the comparisons.
This protocol is similar to the filtered setting\CiteP{FB15k}, with the difference that the rank among entities with tied scores is randomly assigned.

\subsection{KG-BERT}
\FigureW
{\includegraphics[clip, trim=0cm 0cm 0cm 0cm, width=0.95\textwidth ]{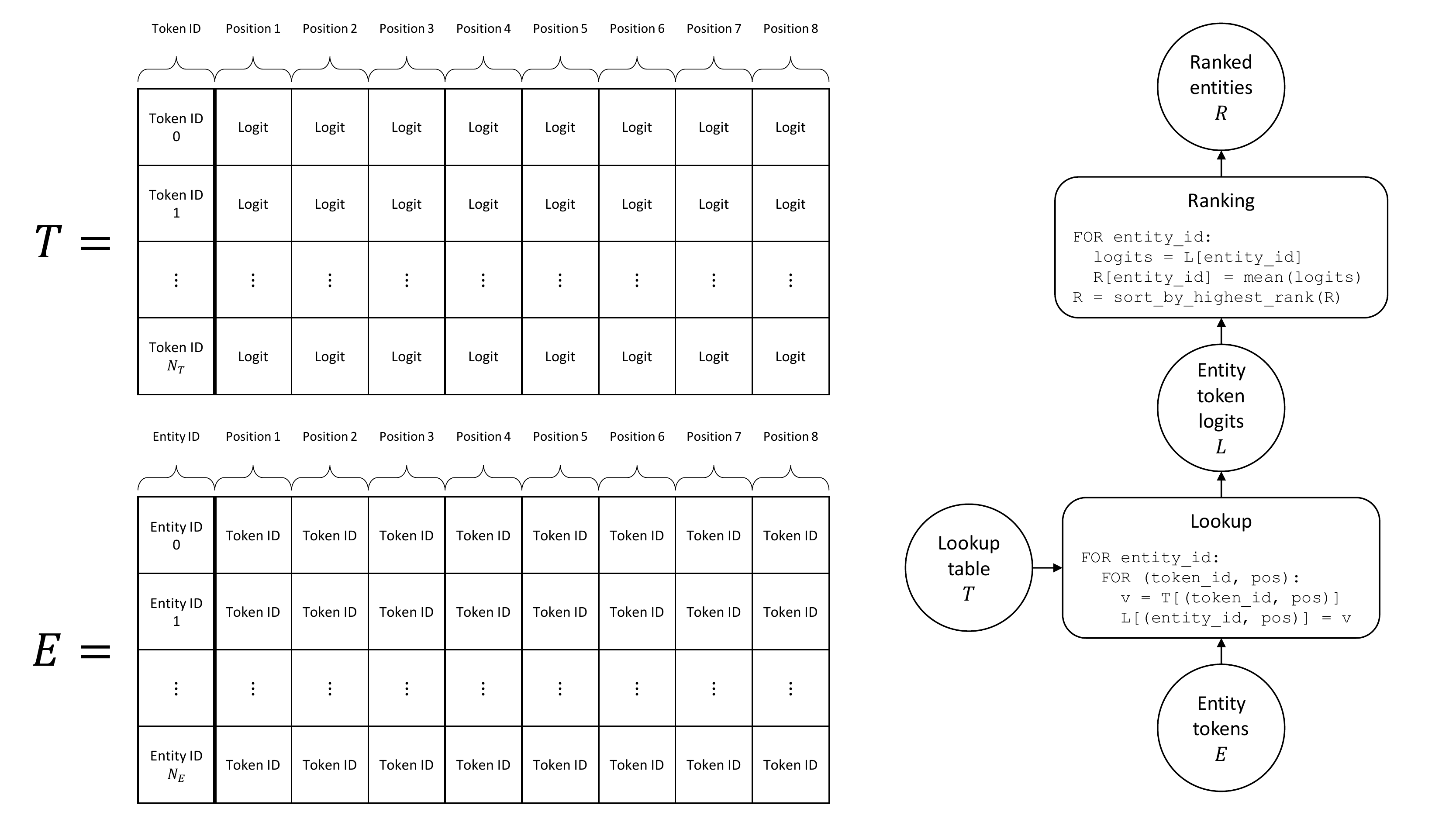}}
{fig:ranking_system}
{Ranking System}
{The figure details the inner workings of the ranking system which uses the lookup table generated by the masked language model to compute the score associated with each possible entity.
The scored entities are then ranked by highest score. \vspace{-.5em}}

% \begin{itemize}
%     \item KG-Bert\CiteP{KG-BERT} is another approach for KB completion based on LM models
%     \item It demonstrates that LM models can be used successfully for a variety of KB tasks
%     \item Approach proposed to link prediction is untractable \RefFigure{fig:inference_time}.
%     \item In contrast, ours is tractable and can generalize to unseen entities.
%     \item Modern KB can have up to millions of entities, KG-BERT cannot scale to hundreds of thousands. Our approach can.
% \end{itemize}

KG-BERT\CiteP{KG-BERT} is an approach to KB tasks based on MLM.
It successfully demonstrates the potential of leveraging those models' internal knowledge on KB tasks.
They train a BERT model to classify whether an individual triplet fed to the model is correct or not. 
% Their approach to link prediction, however, is untractable with larger datasets, scaling in calls to the language model linearly with the total amount of entities to rank.
In essence, they feed every single possible (h, r, ?) and (?, r, t) triplet to the model to obtain all scores to be reranked.
This can result in millions of inference steps on the MLM for a single triplet completion.
In contrast, our approach requires only one inference step through the MLM model for every triplet completion, by generating all logits required to obtain the likelihood of any potential entity. 
A comparison of the evaluation time is pictured in \RefFigure{fig:inference_time}.
Modern KBs can contain millions of entities.
Approaches like KG-BERT cannot scale to hundreds of thousands of entities at evaluation time, having an MLM inference complexity of $O(N)$ where we boast a constant complexity with relation to the number of entities within the KB.

\vspace{-1em}

\Figure
{\includegraphics[clip, trim=0cm 0cm 0cm 0cm, width=0.48\textwidth ]{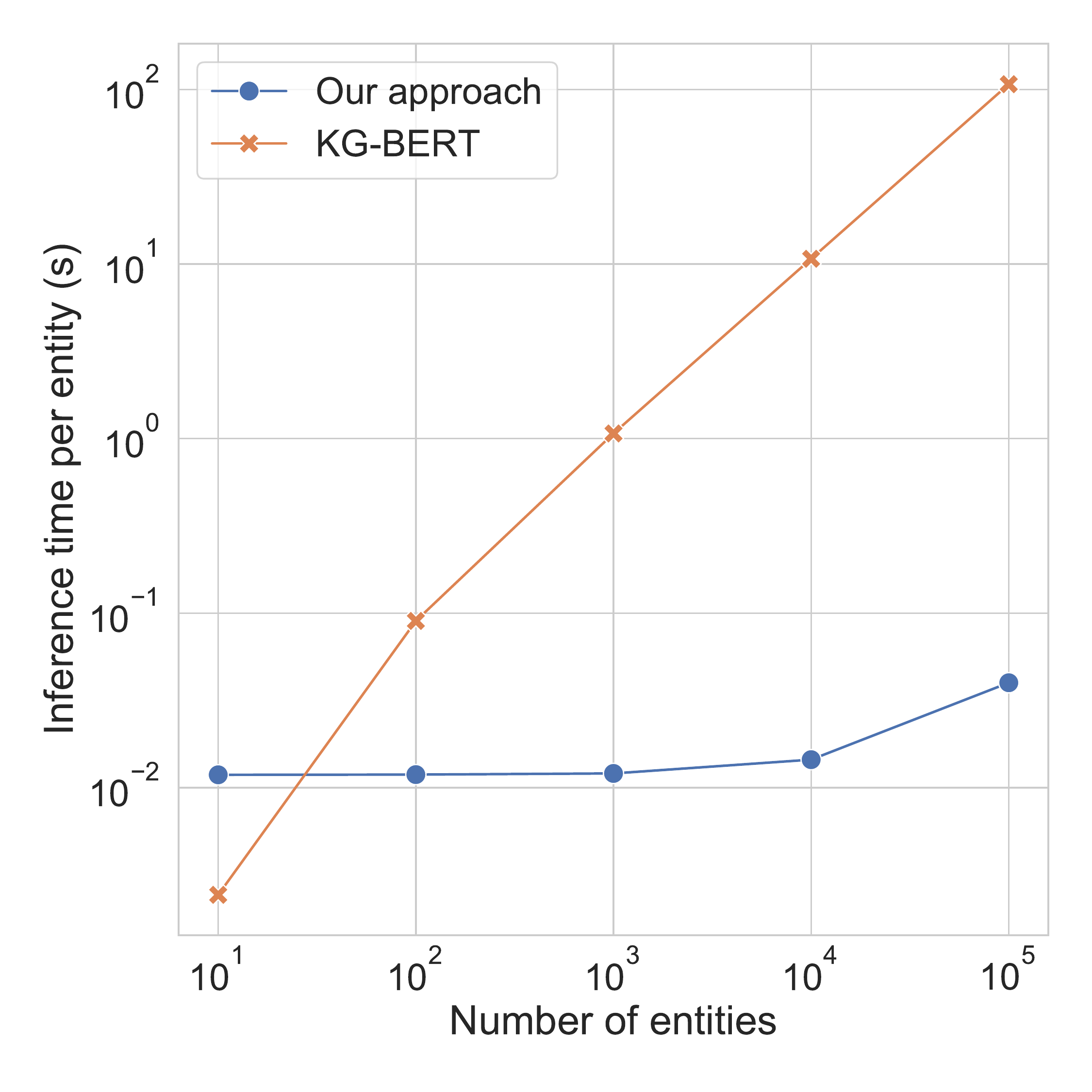}}
{fig:inference_time}
{Approach Inference Time}
{This figure shows the per-entity inference time based on the total number of entities to be re-ranked, of MLMLM and KG-BERT, the most comparable approach. \vspace{-1.5em}}

\section{Methodology}
% This section details the datasets, transformations, models, metrics, and processing used to implement and run our proposed approach, MLMLM.

\subsection{Overview}
Our system performs link prediction.
It uses MLM to generate all possible logits of all tokens required to rebuild all entities, and mean likelihood sampling to rerank all possible entities and perform the task.
It can also be used to sample likelihoods for previously unseen entities.
The system overview is as shown in \RefFigure{fig:system_overview}. 

\vspace{-1em}

\Figure
{\includegraphics[clip, trim=8cm 3.5cm 11cm 1.5cm, width=0.48\textwidth ]{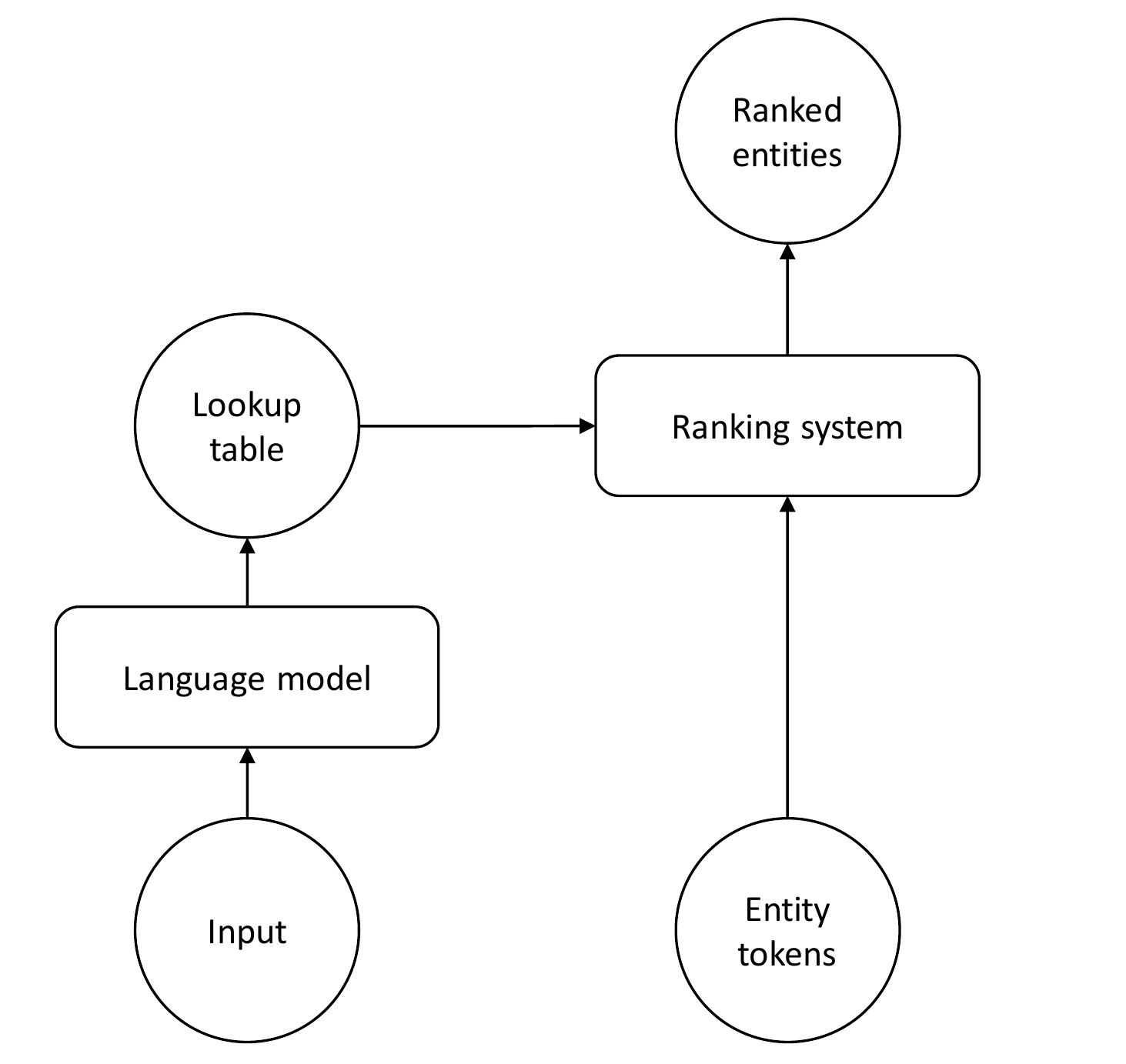}}
{fig:system_overview}
{System Overview}
{This figure presents the system overview.
The inputs are the string representation of the $(h, r, ?)$ or $(?, r, t)$ triplets.
These inputs are then passed to the trained language model to generate a lookup table. 
This lookup table is then used by the ranking system to assign a score to entity tokens based on their likelihood. 
These scores are then finally used to rank the entities, the highest-scoring ones being the best candidates to complete the link.
\vspace{-1.5em}
}

\subsection{Data Pre-processing}\label{sec:dataproc}
% \begin{itemize}
%     \item In our setup, we consider an entity string, a unique string identifying an entity, an entity definition, a description of the entity, and a string representing a relation.
%     \item WN18RR and FB15K-237\CiteP{FB15k-237} are the most commonly used datasets in link prediction
%     \item WN18RR uses word synset, is relatively non dense \RefTable{tab:datasets}.
%     \item For WN18RR, we use the entity definition as the entity string, the synset definition as the entity definition and the cleaned relation as the relation string.
%     \item FB15K-237 is built using entities appearing at least 100 times within FreeBase\CiteP{FB15k}, making it especially dense.
%     \item For FB15k-237, we use the entities name as entity strings, the entity definitions as defined by \CiteT{FBdefinitions} and the cleaned relations.
%     \item WN18RR unseen entities is based on the WN18RR dataset. We randomly sample 10\% of the entities and remove all triplet containing them from the training set. We perform evaluation uniquely on those unseen entities.
% \end{itemize}

The data pre-processing pipeline takes a link prediction dataset and transforms it into a generic format usable by the model.
It is required that both the entity and relations have string representations.
For every entity in the dataset, we extract an entity string, which uniquely identifies the entity, and a definition string, which is a textual description of the given entity.
For every relation, we extract a relation string, which uniquely identifies and describes the relation.

We tokenize all strings through the pretrained RoBERTa tokenizer\CiteP{BPE} and further transform the entity string by adding padding to match the longest tokenized entity within the dataset.
Concretely, in a dataset where the longest entity has a length of 4 token ids, the entity string ``dog'' would be padded to have the representation ``dog~\_~\_~\_'' and the entity string "cat and dog" would have the representation ``cat and dog~\_'' where ``\_'' is the padding token. 
The purpose of this padding is to make the masked representation of all entities the same for the model, therefore letting the model treat all entities in the same manner.

% \subsection{Pre-processing}

% \begin{itemize}
%     \item Second transformation step: from generic format to masked format
%     \item Masking to match the longest entity in the list
% \end{itemize}

\subsection{Model}

\Figure
{\includegraphics[clip, trim=6cm 3cm 6cm 1cm, width=0.48\textwidth ]{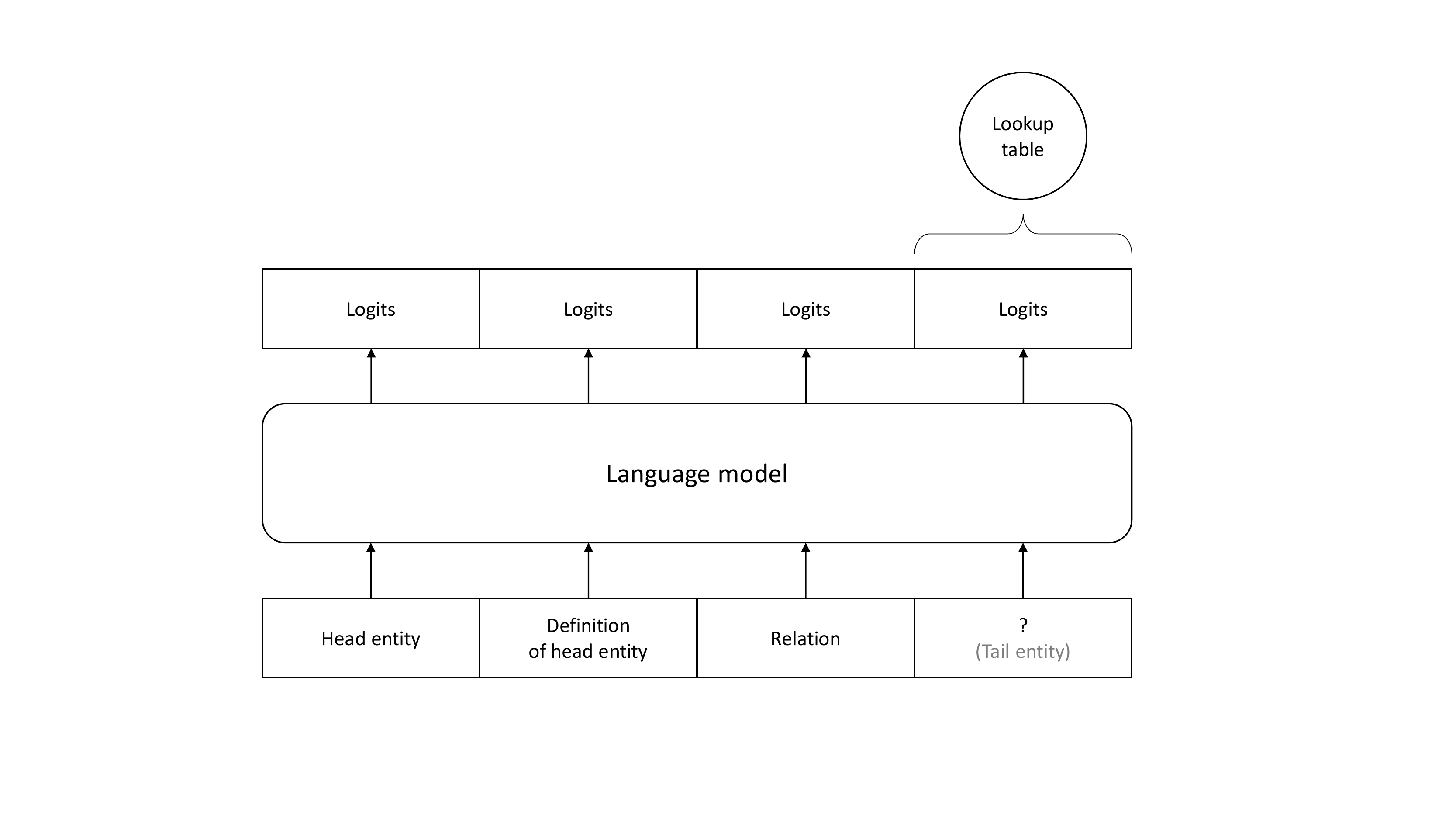}}
{fig:input_predict_tail}
{Lookup Table Generation For Tail Entity Prediction}
{The figure shows how the lookup table for tail entity prediction is generated. A string representation of the head entity and the relation are fed to the masked language model which outputs logits that represent the likelihood of finding each token at each possible position of the tail entity.
\vspace{-1em}
}

% \Figure
% {\includegraphics[clip, trim=6cm 3cm 6cm 1cm, width=0.48\textwidth ]{./res/img/LPMLM_InputPredictHead.pdf}}
% {fig:input_predict_head}
% {Lookup Table Generation For Head Entity Prediction}
% {The figure shows how the lookup table for head entity prediction is generated. A string representation of the tail entity and the relation are fed to the masked language model which outputs logits that represent the likelihood of finding each token at each possible position of the head entity.
% \vspace{-1em}}

% \begin{itemize}
%     \item All experiment run with the pretrained RoBERTa-Large model
%     \item Our approach does a single inference per re-ranking, it is thus acceptable to use a very large model
% \end{itemize}
Our approach uses the RoBERTa-Large model\CiteP{RoBERTa} for all experiments.
It finetunes the pretrained model on the link prediction datasets to generate the logits of the unknown entities.
As our approach does a single call to the model to rerank all possible entities, it is acceptable to use the larger model for better performance.
\RefFigure{fig:input_predict_tail} shows the inference process for tail entity prediction.
The head entity prediction would take as input the head entity mask, the relation, the tail entity and the tail entity definition.
We use the relation string, the known entity string and the entity definition of the known entity string to make the model generate the logits representing the unknown entity string.

\subsection{Ranking System}

\label{sec:datasets}
\TableW{%
\begin{tabular}{ccccc}
    \toprule
    Dataset & Entities & Relations & Mean in-degree & Median in-degree\\
    \midrule
    WN18RR & 40943 & 11 & 2.12 & 1\\
    FB15k-237 & 14541 & 237 & 18.71 & 8\\
    \bottomrule
  \end{tabular}}
{tab:datasets}
{Datasets}

% \begin{itemize}
%     \item Inference through MLM, results in lookup table of logits values
%     \item Entities score obtained through matching the entities token ids to outputs logit values
%     \item Final reranking with randomized setting\CiteP{ReevaluationKB}
% \end{itemize}
The ranking system pictured in \RefFigure{fig:ranking_system} performs link prediction on a given triplet.
The MLM outputs logits for all possible token ids and positions for the missing entity to complete the triplet.
This forms the lookup table~$T$.
The link prediction dataset contains a list of all possible entities.
The token ids forming those entities make up~$E$.
We obtain the entity token logits~$L$ by matching all token ids in~$E$ with their corresponding values in~$T$.
$L$ represents how likely every token of the entity was to be generated by the MLM at that specific position.
% The ranking system performs a lookup using the lookup table~$T$ produced by the language model for the tokens of each possible entity~$E$ to produce the entity token logits~$L$.
The mean likelihood\footnote{Because the length of non-padded tokens is variable, using the mean of the logits is the correct comparison metric for re-ranking.} of each entity is computed by averaging $L$ over \textit{non-padded} token logits\footnote{By far, the token the model sees most is the padding token. Counting it would most likely yield a heavy skew towards shorter entities with more padding.}.
This value is used to determine the ranking of the entity.
It provides a proper comparison between entities of different lengths.

Concretely, in our previous ``cat and dog~\_'' example, we average the outputted logits for the ``cat and dog'' token ids and positions while ignoring the final padded logit.
This averaging is done on all entities in the dataset completing the triplet, yielding the average likelihood assigned by the model to all entities.

Entities are then sorted by highest rank using the randomized setting\CiteP{ReevaluationKB}, meaning that for equal scores the tie-breaking is done randomly, to produce the ordered list of ranked entities~$R$.
We use the filtered setting\CiteP{FB15k} for evaluation and remove corrupted triplets from the list of ranked entities, corrupted triplets being all other known correct triplets.

\section{Experimentation}
\subsection{Datasets}
The two datasets used are WN18RR and FB15K-237\CiteP{FB15k,FB15k-237,WN18RR,wordnet,FreeBase}, two commonly used link prediction benchmarks.
Summary stats for both are shown in \RefTable{tab:datasets}.

WN18RR is a dataset composed of WordNet synsets.
We use the cleaned synset as the entity string.
The synset ``dog.n.01'' would have a string representation of ``dog noun 1'' which should be more interpretable by the model while remaining a unique identifier.
The entity definition is the definition of the entity given by WordNet.
The relation string is the cleaned relation.
The relation ``\_member\_of\_domain\_usage'' would be represented with the string ``member of domain usage''.
Full examples of inputs and outputs are shown in \RefListing{lst:example1} and \RefListing{lst:example2}.

FB15k-237 is composed of triplets found in the now-defunct FreeBase KB, limiting itself to entities appearing in at least 100 triplets.
We use the entity string and definitions as defined in \CiteT{FBdefinitions}. 
We clean the relations to only include the words.

\subsection{Metrics}\label{sec:metrics}
% \begin{itemize}
%     \item We use the filtered setting\CiteP{FB15k} (good answ conflicts)
%     \item A Re-evaluation of Knowledge Graph Completion Methods\CiteP{ReevaluationKB} random evaluation approach. (if rank eq, random)
%     \item Mean Reciprocal Rank, mean precision @ 1, mean precision @ 3, mean precision @ 10 and Mean Rank are the most commonly used metrics and are reported for all experiments.
% \end{itemize}
We use the Mean Reciprocal Rank (MRR) metric to validate our model and select the best model.
For all experiments, we also report the Mean Rank (MR), the Mean Precision at 1 (MP@1), the Mean Precision at 3 (MP@3), and the Mean Precision at 10 (MP@10).

\subsection{Training}

\TableW{%
\begin{tabular}{m{0.3\textwidth}m{0.1\textwidth}m{0.1\textwidth}m{0.1\textwidth}m{0.1\textwidth}m{0.1\textwidth}}
    \toprule
    Approach & MRR $\uparrow$ & MR $\downarrow$ & MP@1 $\uparrow$ & MP@3 $\uparrow$ & MP@10 $\uparrow$\\
    \midrule
    ConvE & 0.444 & 4950 & --- & --- & 0.503\\
    RotatE & 0.473 & 3343 & --- & --- & 0.571\\
    TuckER & 0.461 & 6324 & --- & --- & 0.516\\
    \midrule
    ConvKB & 0.249 & 3433 & --- & --- & 0.524\\
    CapsE & 0.415 & \textbf{718} & --- & --- & 0.559\\
    KBAT & 0.412 & 1921 & --- & --- & 0.554\\
    \midrule
    MLMLM & 
        \shortstack[l]{\textbf{0.5017}\\\textbf{\small$\pm$ 0.0018}} &
        \shortstack[l]{1603\\{\small$\pm$ 26.8184}} &
        \shortstack[l]{{0.4391}\\{\small$\pm$ 0.0020}} &
        \shortstack[l]{{0.5418}\\{\small$\pm$ 0.0028}} &
        \shortstack[l]{\textbf{0.6110}\\\textbf{\small$\pm$ 0.0020}}\\
    \bottomrule
    \multicolumn{6}{p{.8\textwidth}}{\small The results are reported as $<$mean$>$ $\pm$ $<$standard deviation$>$. Results for other models are taken from \CiteT{ReevaluationKB}.}
  \end{tabular}}
{tab:results_WN18}
{WN18RR Results}

%     \item MLM approach, letting the model generate the missing entity
%     \item Challenge of size is dealt with padding, \CodeT{dog} becomes \CodeT{dog\_\_\_\_\_\_\_} to match the length 8.
%     \item Model takes head (or tail) entity and relation in input, as text, as well as masks, and outputs the tail (or head) entity
%     \item We rerank all entities by averaging their token logits outputted by the model, letting us compare all entities with a single inference
%     \item MLM training is used

The training setup is a modified MLM training, where we let the model generate the missing entity.
The previously mentioned padding lets us deal with the generation of entities of varying sizes.
The input fed to the model for tail entity prediction, pictured in \RefFigure{fig:input_predict_tail}, consists of the concatenated token ids of the head entity, the head entity definition, the relation and the tail entity mask.
The model will then generate, in the place of the mask, the missing entity.
The input fed to the model for head entity prediction is similar.
An example of the input for head entity prediction is found in \RefListing{lst:example1} and an example for tail entity prediction is found in \RefListing{lst:example2}.

We use the categorical cross-entropy loss to train the language model.
The loss only depends on the non-padded token of the generated entity, ignoring all other outputs.
The target is the actual entity completing the triplet, aligned with the mask in the input.
We retain the model with the best validation MRR.
All experiments are run for 5 random seeds and the mean and standard deviation of the results are reported.

For all experiments, we use the hyperparameters and training setup described in \CiteT{RoBERTa} and shown in \RefTable{tab:hparams}, with a total of 10 epochs for the FB15k-237 dataset and 25 epochs for the WN18RR dataset.

\Table{%
\begin{tabular}{m{0.28\textwidth}l}
    \toprule
    Hyperparameter & Value\\
    \midrule
    Max sequence length & $512$\\
    Batch size & $32$\\
    Learning rate & $2 \times 10^{-5}$\\
    Weight decay & $0.1$\\
    Gradient Norm & $1.0$\\
    \bottomrule
  \end{tabular}}
{tab:hparams}
{All experiments hyperparameters.}

\subsection{Unseen Entities}

A secondary version of the dataset is made to test the generalization capacity of our methodology to unseen entities.
For both datasets, we start by randomly sampling 5\% of the entities for the validation entities and 5\% of the entities for the testing entities.
Our training set consists of all triplets not containing any of the validation or testing entities.
Our validation set consists of all triplets containing the validation entities.
Finally, our test set consists of all triplets containing the test entities, but not containing any of the validation entities.
The training is done in the same fashion.
The validation and testing are only done on entities present in the validation or test entity list.
If the tail entity is the one present in the test entity list, we will complete the link $(h, r, ?)$ and not the link $(?, r, t)$.
The reported results are therefore only on the performance of previously unseen entities in the KB.
The validation and test set are rebuilt for every random seed, to evaluate our approach on a wider array of unseen entities.

\begin{lstlisting}[float=*,basicstyle=\ttfamily\small,breaklines=true,label={lst:example1},caption={
Example of an error of the model on WN18RR. Shown are the top 5 ranked entities by the model with the score assigned to them.
The correct answer, \texttt{matchmaker noun 1}, was ranked 14,108 by the system.
}]
Prompt : <s><mask><mask><mask><mask><mask><mask><mask><mask>hypernym mediator noun 1 a negotiator who acts as a link between parties</s><pad><pad><pad><pad>
Correct answer : matchmaker noun 1<pad><pad><pad><pad> 	 Answer rank 14108
Rank 1	 Score 32.0242	 : interpreter noun 2<pad><pad><pad>
Rank 2	 Score 32.0103	 : harmonizer noun 1<pad><pad><pad>
Rank 3	 Score 31.8889	 : diplomat noun 1<pad><pad><pad>
Rank 4	 Score 31.8286	 : interpreter noun 4<pad><pad><pad>
Rank 5	 Score 31.1707	 : conciliation noun 2<pad><pad><pad>
\end{lstlisting}

\begin{lstlisting}[float=*,basicstyle=\ttfamily\small,breaklines=true,label={lst:example2},caption={
Example of a disambiguation error of the model on WN18RR. Shown are the top 5 ranked entities by the model with the score assigned to them.
The correct answer, \texttt{aid noun 3}, was ranked second by the system, after \texttt{aid noun 1}.
}]
Prompt : <s>grant noun 1 any monetary aid hypernym<mask><mask><mask><mask><mask><mask><mask><mask></s><pad><pad><pad><pad>
Correct answer : aid noun 3<pad><pad><pad><pad><pad> 	 Answer rank 2
Rank 1	 Score 33.7597	 : aid noun 1<pad><pad><pad><pad><pad>
Rank 2	 Score 33.5948	 : aid noun 3<pad><pad><pad><pad><pad>
Rank 3	 Score 32.7605	 : aid noun 2<pad><pad><pad><pad><pad>
Rank 4	 Score 31.4054	 : interest noun 1<pad><pad><pad><pad><pad>
Rank 5	 Score 31.3839	 : need noun 1<pad><pad><pad><pad><pad>
\end{lstlisting}

\section{Results and Analysis}
\subsection{WN18RR}

We achieve SotA results on the WN18RR dataset on all tested metrics with the exception of MR, shown in \RefTable{tab:results_WN18}.
% This implies the model generally performs well, but tends to have more complete misses than the CapsE model.
The WN18RR dataset is sparse in terms of relations, see \RefTable{tab:datasets}.
This sparseness lends itself naturally to leveraging a pretrained model, since the amount of information that can be extracted from the dataset on any given entity is limited, which makes outside information all the more valuable.
% Hard to defend the style of writing since the models are trained a lot more on FB15k style string than on formal language like WN18RR, WN18RR is actually more out of domain than FB15k is.

We can observe a large discrepancy between the MP@1 and MP@3 metrics, implying that the model will have the correct answer in its top 3 much more often than within its top 1. 
This could be explained by an issue of disambiguation in the name of the entity.
While approaches using entity embeddings~\CiteP{EMB1, EMB2, EMB3} will have no issue separating the synsets \texttt{dog.n.01} and \texttt{dog.n.03} as meaning respectively ``a member of the genus Canis [...]'' and ``informal term for a man'', our model will have to discern between those two meanings only by the digit appended to the name.
It is probable that the model is often confused about whether it should generate \texttt{dog~noun~1} or \texttt{dog~noun~3}, having only the final digit to differentiate both of them.
An example of such an error is shown in \RefListing{lst:example2}, where the model confuses \texttt{aid.n.01} and \texttt{aid.n.03}.
Follow up work on better representations for entity names could yield stronger results.
\TableW{%
\begin{tabular}{m{0.3\textwidth}m{0.1\textwidth}m{0.1\textwidth}m{0.1\textwidth}m{0.1\textwidth}m{0.1\textwidth}}
    \toprule
    Approach & MRR $\uparrow$ & MR $\downarrow$ & MP@1 $\uparrow$ & MP@3 $\uparrow$ & MP@10 $\uparrow$\\
    \midrule
    ConvE & 0.324 & 285 & --- & --- & 0.501\\
    RotatE & 0.336 & 178 & --- & --- & 0.530\\
    TuckER & \textbf{0.353} & \textbf{162} & --- & --- & \textbf{0.536}\\
    \midrule
    ConvKB & 0.243 & 309 & --- & --- & 0.421\\
    CapsE & 0.150 & 403 & --- & --- & 0.356\\
    KBAT & 0.157 & 270 & --- & --- & 0.331\\
    \midrule
    MLMLM & 
        \shortstack[l]{{0.2591}\\{\small$\pm$ 0.0017}} &
        \shortstack[l]{{411.23}\\{\small$\pm$ 0.0014}} &
        \shortstack[l]{{0.1871}\\{\small$\pm$ 0.0028}} &
        \shortstack[l]{{0.2820}\\{\small$\pm$ 0.0017}} &
        \shortstack[l]{{0.4026}\\{\small$\pm$ 2.9313}}\\
    \bottomrule
    \multicolumn{6}{p{.8\textwidth}}{\small The results are reported as $<$mean$>$ $\pm$ $<$standard deviation$>$. Results for other models are taken from \CiteT{ReevaluationKB}.}

  \end{tabular}}
{tab:results_FB15k}
{FB15k-237 Results}

We performed some quantitative and qualitative error analysis to understand some of the remaining shortcomings of our approach.
It seems like our model generally has a much easier time predicting the tail entity than the head entity, having an MRR of 0.6015 on tail entities and an MRR of 0.4009 on head entities.
By observing the instances where our model gives the worst rank to the correct answer, we can understand why.
A large number of those cases are hypernyms on the head entity.
The definition of a hypernym is as follows: ``A hypernym of something is its superordinate term: if X is a hypernym of Y, then all Y are X.''\CiteP{wordnet}.
An example of a hypernym relationship would be: ``animal is an hypernym of dog, since all dogs are animals.''
Correctly ranking all possibilities for ``X is an hypernym of dog.'' seems easier for the model to do than correctly ranking all possibilities for ``Animal is an hypernym of Y.''.
An example of such failure is shown in \RefListing{lst:example1}, where we look for the hypernym of the term \texttt{mediator}.
It is clear that the model understands the concept and outputs plausible answers in its top 5.
A large amount of the model's severe failure cases are similar to this one, where the model will output a plausible hypernym of the tail entity, while completely missing the targeted hypernym.

% \begin{itemize}
%     \item SotA
%     \item WN18RR being a very sparse graph, it lends itself naturally to leveraging a pretrained model. 
%     \item Peraphs some disambiguation issues (CatNoun1 vs CatNoun2) explain the relatively weak MP1 compared to MP10 and even MP3.
% \end{itemize}

\subsection{FB15K-237}
\TableW{%
\begin{tabular}{m{0.34\textwidth}m{0.08\textwidth}m{0.1\textwidth}m{0.1\textwidth}m{0.1\textwidth}m{0.1\textwidth}}
    \toprule
    Approach & MRR $\uparrow$ & MR $\downarrow$ & MP@1 $\uparrow$ & MP@3 $\uparrow$ & MP@10 $\uparrow$\\
    \midrule 
    %TODO add std and mean for all experiments
    Random baseline & 
        \shortstack[l]{{0.0003}\\{\small$\pm$ 0.00007}} &
        \shortstack[l]{{20541.91}\\{\small$\pm$ 87.88}} &
        \shortstack[l]{{0.00002}\\{\small$\pm$ 0.00004}} &
        \shortstack[l]{{0.00002}\\{\small$\pm$ 0.00004}} &
        \shortstack[l]{{0.00026}\\{\small$\pm$ 0.00008}}\\
    Non-finetuned RoBERTa & 
        \shortstack[l]{{0.0273}\\{\small$\pm$ 0.0005}} &
        \shortstack[l]{{10130.35}\\{\small$\pm$ 187.61}} &
        \shortstack[l]{{0.0154}\\{\small$\pm$ 0.0007}} &
        \shortstack[l]{{0.0295}\\{\small$\pm$ 0.0011}} &
        \shortstack[l]{{0.0492}\\{\small$\pm$ 0.0019}}\\
    MLMLM & 
        \shortstack[l]{{0.1842}\\{\small$\pm$ 0.0266}} &
        \shortstack[l]{{3761.50}\\{\small$\pm$ 255.4437}} &
        \shortstack[l]{{0.1416}\\{\small$\pm$ 0.0081}} &
        \shortstack[l]{{0.2175}\\{\small$\pm$ 0.0119}} &
        \shortstack[l]{{0.2939}\\{\small$\pm$ 0.0088}}\\
    \bottomrule
    \multicolumn{6}{p{.8\textwidth}}{\small The results are reported as $<$mean$>$ $\pm$ $<$standard deviation$>$.}

  \end{tabular}}
{tab:results3}
{WN18RR Unseen Entities Result}
\TableW{%
\begin{tabular}{m{0.34\textwidth}m{0.08\textwidth}m{0.1\textwidth}m{0.1\textwidth}m{0.1\textwidth}m{0.1\textwidth}}
    \toprule
    Approach & MRR $\uparrow$ & MR $\downarrow$ & MP@1 $\uparrow$ & MP@3 $\uparrow$ & MP@10 $\uparrow$\\
    \midrule 
    Random baseline & 
        \shortstack[l]{{0.0007}\\{\small$\pm$ 0.00011}} &
        \shortstack[l]{{7065.95}\\{\small$\pm$ 12.29}} &
        \shortstack[l]{{0.00006}\\{\small$\pm$ 0.00012}} &
        \shortstack[l]{{0.00026}\\{\small$\pm$ 0.00008}} &
        \shortstack[l]{{0.00074}\\{\small$\pm$ 0.00013}}\\
    Non-finetuned RoBERTa & 
        \shortstack[l]{{0.0115}\\{\small$\pm$ 0.0028}} &
        \shortstack[l]{{4870.56}\\{\small$\pm$ 437.03}} &
        \shortstack[l]{{0.0060}\\{\small$\pm$ 0.0013}} &
        \shortstack[l]{{0.0101}\\{\small$\pm$ 0.0016}} &
        \shortstack[l]{{0.0190}\\{\small$\pm$ 0.0069}}\\
    MLMLM & 
        \shortstack[l]{{0.0694}\\{\small$\pm$ 0.01823}} &
        \shortstack[l]{{2057.61}\\{\small$\pm$ 293.94}} &
        \shortstack[l]{{0.0258}\\{\small$\pm$ 0.0019}} &
        \shortstack[l]{{0.0768}\\{\small$\pm$ 0.0400}} &
        \shortstack[l]{{0.1499}\\{\small$\pm$ 0.0410}}\\
    \bottomrule
    \multicolumn{6}{p{.8\textwidth}}{\small The results are reported as $<$mean$>$ $\pm$ $<$standard deviation$>$.}
  \end{tabular}}
{tab:results4}
{FB15k-237 Unseen Entities Result}

The results on FB15k-237 shown in \RefTable{tab:results_FB15k} are, comparatively to the results obtained on WN18RR, fairly weak.
FB15k-237 is very dense and contains a lot more training examples than the WN18RR dataset for a smaller amount of entities.
Thus, non-pretrained models have way more examples to learn from in the dataset, which makes the learned information of pretrained models comparatively less impactful.
This implies, non-surprisingly, that our approach heavily relies on the pre-training of the model and that it is less adept than other specialized approaches at learning from dense link prediction datasets.

However, FB15k-237 is an especially dense section of the FreeBase dataset, being composed of only entities containing a minimum of 100 relations, and is thus not representative of the KB as a whole.
In practice, KB completion will often be used on entities rarely or never seen within the KB.
While our FB15k-237 results are not SotA when compared to all approaches, the MRR however compares favorably to all other non entity-embedding approaches on the randomized setting.

\subsection{Unknown Entities Experiments}
We demonstrate the capacity of our approach to generalize to unknown entities.
Results for the WN18RR and the FB15k-237 datasets are shown in \RefTable{tab:results3} and \RefTable{tab:results4}.

For baselines, we use a random baseline, reranking the entities randomly, as well as a non-finetuned RoBERTa-large model, that simply generates the entity tokens without being finetuned on the dataset first.
We can notice that while our approach outperforms a non-finetuned benchmark, the non-finetuned RoBERTa model still far outperforms the random baseline, supporting some of the findings of \CiteT{LMKB} in the capacity of MLM to perform unsupervised link prediction.

It is to be noted that the high standard deviation of the results in this set of experiments comes from the fact that the validation and test entities are resampled with a different random seed on every run, yielding more variability in the results.

We are unaware of other approaches that can generalize to unknown entities of arbitrary size in the task of link prediction.
We believe that leveraging MLMs could eventually lead to automatically populating KBs with new entities, as new knowledge and new facts are created and added to the web.

\subsection{Limitations}
MLMLM comes with several limitations.
Our approach to padding limits the size of an unknown entity to the size of the longest known entity. 
While it is likely to not be limiting in practice, it is still a weakness of our approach to sampling.
The model size can be very prohibitive and specialized hardware such as GPUs is required to run it in a timely fashion.
The approach however remains tractable as it can provide likelihoods for all possible entities in a single inference call.
Compared to entity-embedding based methods, our approach needs additional information in the form of meaningful string representations for both entities and relations.
Entity disambiguation is also a limiting factor that does not affect other approaches. 

\section{Conclusion}
We have developed a methodology for training masked language models to perform link prediction. 
By leveraging the natural language understanding abilities of these models as well as the factual knowledge embedded within their weights, we have achieved a tractable approach to link prediction that yields state of the art results on a standard benchmark and the best non entity-embedding based results on another.
We have also demonstrated the ability of our model to perform link prediction of previously unseen entities, making our approach suitable to introduce new entities to knowledge bases.
More generally, we have introduced an approach to sampling text from a masked language model of varying lengths, which can have a wider use case.
\clearpage

\nocite{HuggingFace}
\nocite{pytorch}
\bibliography{anthology,emnlp2020}
\bibliographystyle{acl_natbib}

% \appendix

% \section{Appendices}
% \label{sec:appendix}
% Appendices are material that can be read, and include lemmas, formulas, proofs, and tables that are not critical to the reading and understanding of the paper. 
% Appendices should be \textbf{uploaded as supplementary material} when submitting the paper for review.
% Upon acceptance, the appendices come after the references, as shown here.

% \paragraph{\LaTeX-specific details:}
% Use {\small\verb|\appendix|} before any appendix section to switch the section numbering over to letters.

\end{document}